\documentclass{article}
\usepackage{spconf,amsmath,graphicx}

\usepackage{enumitem}
\setlist{nosep, leftmargin=14pt}

\usepackage{tabularx}
\usepackage{xcolor}
\usepackage{multirow}
\usepackage{amssymb} 

\newcommand{\hgl}[1]{\textcolor{blue}{#1}}
\definecolor{commentred}{RGB}{240,0,0} 
\definecolor{commentgreen}{RGB}{0,200,0} 
\def\inc#1{\textcolor{commentgreen} {\scriptsize $\uparrow$ \hspace{-1.2mm}  #1} \color{black}}
\def\dec#1{\textcolor{commentred} {\scriptsize $\downarrow$ \hspace{-1.2mm}  #1} \color{black}}
\newcommand{\mname}{UKAST}

\usepackage{pifont}
\newcommand{\xmark}{\ding{55}} 

\begin{document}
\title{When Swin Transformer meets KANs: An Improved Transformer Architecture for Medical Image Segmentation \thanks{Published in the Proceedings of IEEE ISBI 2026.
}}
\name{
Nishchal Sapkota$^{\dagger}$
\hspace{-0mm}
\quad 
Haoyan Shi$^{\dagger}$
\hspace{-0mm}
\quad 
Yejia Zhang$^{\dagger}$ \vspace{-0.48cm}
}

\address{
\emph{Xianshi Ma}$^{\dagger}$
\hspace{-0mm}
\quad 
\emph{Bofang Zheng}$^{\dagger}$
\hspace{-0mm}
\quad
\emph{Fabian Vazquez}$^{\star}$
\hspace{-0mm}
\quad
\emph{Pengfei Gu}$^{\star}$
\hspace{-0mm}
\quad
\emph{Danny Z. Chen}$^{\dagger}$ \vspace{2mm} \\
$^{\dagger}$ Department of Computer Science and Engineering,
University of Notre Dame,
IN 46556, USA \\
$^{\star}$ Department of Computer Science, University of Texas Rio Grande Valley, Edinburg, TX 78539
}

%
%
%

%
\maketitle
\begin{abstract}
Medical image segmentation is critical for accurate diagnostics and treatment planning, but remains challenging due to complex anatomical structures and limited annotated training data. 
CNN-based segmentation methods excel at local feature extraction, but struggle with modeling long-range dependencies. Transformers, on the other hand, capture global context more effectively, but are inherently data-hungry and computationally expensive.
In this work, we introduce \mname, a U-Net like architecture that integrates rational-function based Kolmogorov–Arnold Networks (KANs) into Swin Transformer encoders. By leveraging rational base functions and Group Rational KANs (GR-KANs) from the Kolmogorov–Arnold Transformer (KAT), our architecture addresses the inefficiencies of vanilla spline-based KANs, yielding a more expressive and data-efficient framework with reduced FLOPs and only a very small increase in parameter count compared to SwinUNETR.
\mname{} achieves state-of-the-art performance on four diverse 2D and 3D medical image segmentation benchmarks, consistently surpassing both CNN- and Transformer-based baselines.
Notably, it attains superior accuracy in data-scarce settings, alleviating the data-hungry limitations of standard Vision Transformers. These results show the potential of KAN-enhanced Transformers to advance data-efficient medical image segmentation. Code is available at: \hgl{https://github.com/nsapkota417/UKAST}
\end{abstract}

\section{Introduction}
\label{sec:intro}

Medical image segmentation is essential for delineating anatomical structures and pathological regions, supporting diagnosis, quantitative analysis, and treatment planning.
Achieving high segmentation accuracy is crucial as even small errors could affect clinical decisions. 
However, medical image datasets are often constrained by expensive expert annotations and privacy restrictions, making data efficiency a key challenge. 
Convolutional Neural Network (CNN)–based models such as U-Net \cite{unet} and its variants \cite{isensee2021nnuNet, valanarasu2022unext, milletari2016vnet} have been widely adopted, exploiting locality and translation invariance. But, their limited receptive fields restrict long-range dependency modeling, which is vital for segmenting objects of varying shapes and sizes. 
Vision Transformers (ViT)~\cite{vit} and 
ViT-based models~\cite{hatamizadeh2022unetr, chen2021transunet, sapkota2024conunetr} addressed this with self-attention \cite{transformer} to capture global relationships and improve context modeling. 
Still, the vanilla ViT scales poorly for dense prediction (as tokens grow quadratically with resolution) and limits its practicality for high-resolution 
images. 
The Swin Transformer (SwinT) \cite{liu2021swin} and SwinT-based image segmentation models \cite{cao2022swinUnet,hatamizadeh2021swinUnetr, tangSelfSupervisedPreTrainingSwin2022, he2023swinunetrV2, sapkota2025universal} mitigated this issue through hierarchical shifted windows to reduce computation costs to near-linear while preserving cross-window context. 
By extracting multi-scale features and balancing local and global context, SwinT-based models have proven highly effective for medical image dense prediction tasks such as segmentation.
\begin{figure*}
  \centering
  \includegraphics[width=0.89\linewidth]{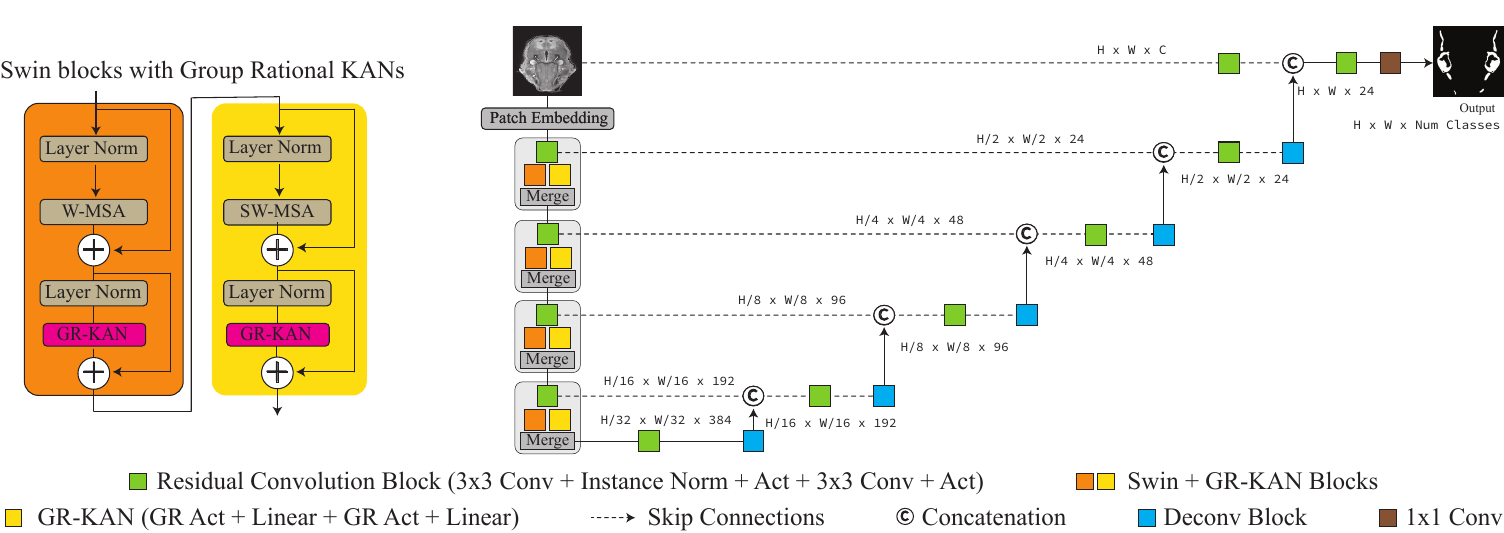}
  \caption{\small{
  The architecture of our proposed \mname{} model consists of SwinT blocks infused with Group Rational KANs, along with CNN decoders connected via hierarchical skip connections.}}
  \label{fig:Model}
\end{figure*}

In ViT and SwinT, Multi-Layer Perceptrons (MLPs) serve as the default Feed-Forward Networks (FFNs) 
to introduce non-linear transformations and channel mixing. 
However, their reliance on fixed activations such as GELU or ReLU limits their ability to capture complex functional relationships in medical images, and most of their variants still model patterns largely in a linear and opaque manner. 
Kolmogorov–Arnold Networks (KANs) \cite{liu2024kan} address these limitations by replacing static activations with learnable functional expansions, offering greater expressiveness and interpretability. By parameterizing the base functions along each dimension, KANs bridge the gap between a network’s physical attributes and empirical performance. In computer vision, they enhance representation learning by capturing richer non-linear dependencies \cite{bodner2024convolutionalCKAN}, and in medical image analysis, they provide a path toward improved data efficiency and finer anatomical detail modeling. Recent work, such as U-KAN \cite{li2025ukan}, has shown the potential of incorporating KAN layers into U-Net type architectures, enhancing expressiveness and data efficiency for medical image segmentation. 
Although effective, UKAN's reliance on convolutions limits the modeling of long-range dependencies, while its design of using tokenized-KANs only at the bottleneck stage limits their expressive power.



In this work, we explore integrating KANs into Transformer encoders for medical image segmentation, building a more accurate and data-efficient architecture. This design combines the complementary strengths of both components: self-attention in Transformers to capture long-range 
context essential for complex anatomical structures, and KANs to enrich the feed-forward pathways with flexible and interpretable non-linear approximations for better representation capacity and data efficiency. 
Transformers with KANs have been investigated in recent work \cite{wang2025swinkan, rifa2025swinKAT} using hybrid MLP–KAN or KAN-convolution designs, while TransUKAN \cite{wu2024transukan} employs CNNs with a KAN bottleneck. In contrast, we integrate KANs directly as feed-forward layers in a standalone SwinT encoder, making ours the first pure Transformer-based encoder with KANs for medical image segmentation. 

It is challenging to scale KANs in deep architectures, as vanilla spline-based KANs are computationally inefficient, memory-intensive, and difficult to train~\cite{liu2024kan}. 
To address this, we adopt rational base functions and the Group Rational KAN (GR-KAN) design from KAT~\cite{yang2024kolmogorovKAT} to build a more robust Swin Transformer–based architecture.
Our proposed model, \textbf{\mname}
(\underline{U}-shaped segmentation architecture combining \underline{KA}Ns with \underline{S}win \underline{T}ransformers), mitigates the computation and training inefficiencies of vanilla KANs, yielding a data-efficient segmentation framework with lower FLOPs and comparable parameters as to state-of-the-art (SOTA) SwinUNETR. Our \mname{} achieves SOTA performance across diverse modalities of medical image segmentation benchmarks, consistently surpassing both CNN- and Transformer-based baselines. 
It delivers improved segmentation accuracy in two 2D and even greater improvements in two 3D datasets with approximately the same computational costs in both FLOPs and model parameters. 
We demonstrate that UKAST's data efficiency outperforms that of ViTs and SwinTs across various data-scarce settings in both 2D and 3D tasks.

Our main contributions are four-fold: We 
(1) propose a novel architecture, \mname, by integrating rational-function-based KANs into Swin Transformers by overcoming the fundamental limitations of KANs, 
(2) achieve SOTA segmentation results on both 2D and 3D benchmarks across various data modalities,
(3) systematically evaluate Transformers with KANs for medical image segmentation across different data scales, and
(4) demonstrate strong data efficiency of our model under limited annotations.

\begin{table*}[h]
\centering
\resizebox{0.5\textwidth}{!}{
\begin{tabular}{r|r|ll|ll}
\hline
\hline
\textbf{Model} & \textbf{\# Params.} & \multicolumn{2}{c|}{\textbf{2D}}
& \multicolumn{2}{c}{\textbf{3D}} \\
\cline{3-6}
& & \textbf{Kvasir} & \textbf{ISIC} & \textbf{BCV} & \textbf{MMWHS} \\
\hline
\hline
U-Net  & 2.6 M & 71.8 & 77.3 & 59.3 & 71.9 \\
U-Net v2  & 3.7 M & 82.2 & 74.9 & 70.0 & 72.0 \\
UNeXt  & 1.5 M & \textbf{82.3} & 79.3 & 61.9 & 73.4 \\
U-KAN  & 6.3 M & 81.7 & 76.3 & 62.6 & 73.6 \\
\hline
UNETR  & 8.3 M & 63.7 & 74.7 & 52.7 & 70.3 \\
\hgl{UKAT (ours)} & 4.8 M & 70.0 \inc{6.3} & 75.3 \inc{0.6}
& 55.0 \inc{2.3} & 78.0 \inc{7.7} \\
\hline
SwinUNETR + RC 
& 7.2 M & 81.9 & 78.9 & 68.9 & 80.4 \\
\hgl{UKAST (ours)} & 7.2 M & 81.7 \dec{0.2} &
\textbf{79.9} \inc{1.0} & \textbf{71.2} \inc{2.3} &
\textbf{80.8} \inc{0.4} \\
\hline
\hline
\end{tabular}
}
\caption{\small{\textbf{Comparison with SOTA architectures}: Dice scores with 100\% training data are reported.}}
\label{tab:main}
\end{table*}

\vspace{-5mm}
\section{Methodology}
\label{sec:method}

\begin{table*}
\centering
\resizebox{0.7\textwidth}{!}{
\begin{tabular}{r|llll|llll}
\hline
\hline
\multicolumn{1}{r|}{\textbf{Model}} &
\multicolumn{4}{c|}{\textbf{ISIC (2D)}} &
\multicolumn{4}{c}{\textbf{BCV (3D)}} \\
\cline{2-9}
  & \textbf{10\%} & \textbf{25\%} & \textbf{50\%} & \textbf{100\%} & \textbf{10\%} & \textbf{25\%} & \textbf{50\%} & \textbf{100\%} \\
\hline
\hline
U-Net                  & 70.4 & 72.0 & 74.8 & 77.3  & 50.3 & 50.9 & 52.0   & 59.3  \\
U-Net v2               & 65.5 & 68.4  & 71.7 & 74.9 & 49.0 & 53.6 & 54.5 & 70.0 \\
UNeXt                 & 67.8 & 70.3  & 72.8 & 79.3 & 47.2 & 47.5 & 48.9 & 61.9 \\
U-KAN                  & 68.7 & 70.3  & 73.0 & 76.3 & 50.7 & 51.6 & 52.7 & 62.6 \\
\hline
UNETR                 & 68.6 & 70.4 
& 71.9 & 74.7 & 33.1 & 33.7 & 34.5  & 52.7  \\
\hgl{UKAT (ours)}   & 69.4 \inc{0.8} & 71.2 \inc{0.8}  & 72.9 \inc{1.0}   & 75.3 \inc{0.6} 
& 45.1 \inc{12.0} & 49.6 \inc{15.9} & 50.6 \inc{16.1} &55.0 \inc{2.3}  \\
\hline
SwinUNETR + RC    
& 69.8 & 71.8  & 74.3 & 78.9
& 60.1 & 62.7 & 63.1  & 68.9  \\
\hgl{UKAST (ours)}  
& \textbf{71.4} \inc{1.6} & \textbf{72.1} \inc{0.3}  & \textbf{75.5} \inc{1.2} & \textbf{79.9} \inc{1.0} 
& \textbf{63.9} \inc{3.8} & \textbf{64.6} \inc{4.9} & \textbf{67.2} \inc{4.1}  & \textbf{71.2} \inc{2.3} \\
\hline
\hline
\end{tabular}}
\caption{\small{\textbf{Data efficacy results}: Dice scores with different amounts of training data reported.}}
\label{tab:data-efficacy}
\end{table*}

\vspace{-3mm}
\subsection{\mname}
Our model, \mname, adopts a U-shaped architecture that integrates Kolmogorov–Arnold Network with Swin Transformer encoders along with CNN decoders connected through hierarchical skip connections (see Fig.~\ref{fig:Model}).

Given an input 2D medical image of size $H \times W$ with $C$ number of channels, 
\(\mathcal{X} \in \mathbb{R}^{C \times H \times W}\) (for 3D volumes, we operate on their 2D slices), 
we divide \(\mathcal{X}\) into non-overlapping patches of size \(H'\times W'\) each. 
Every patch is projected into a \(C'\)-dimensional embedding space via a linear layer, forming a sequence of input tokens. The input tokens are partitioned into local windows of size \(M \times M\) each in the UKAST encoder, which comprises hierarchical Swin Transformer blocks with Group-KAN feed-forward layers using rational base functions. 
Within each window, self-attention (W-MSA) is computed to capture local dependencies, where residual convolutions (RC) \cite{he2023swinunetrV2} are also added. To enable cross-window interactions, the subsequent layer applies shifted window attention (SW-MSA), balancing local modeling with global context aggregation. 
This hierarchical design progressively extracts multi-scale features, encoding both fine-grained details and broader structural information, as follows. 
{\small{
\[
\begin{aligned}
v^{(s)}_0      &= \text{RC}\big(z^{(s)}_{\text{in}}\big), & &\text{(res. conv. proj. of stage-$s$)} \\
\hat z^{(s)}_1 &= \text{W-MSA}\big(\text{LN}(v^{(s)}_0)\big) + v^{(s)}_0, & &\text{(windowed self-attention)} \\
z^{(s)}_1      &= \text{GR-KAN}\big(\text{LN}(\hat z^{(s)}_1)\big) + \hat z^{(s)}_1, & &\text{(feed-forward update)} \\
\hat z^{(s)}_2 &= \text{SW-MSA}\big(\text{LN}(z^{(s)}_1)\big) + z^{(s)}_1, & &\text{(shifted window attention)} \\
z^{(s)}_{\text{out}} &= \text{GR-KAN}\big(\text{LN}(\hat z^{(s)}_2)\big) + \hat z^{(s)}_2. & &\text{(final stage-$s$ output)}
\end{aligned}
\]
}}
Here, \(z^{(s)}_{\text{in}}\) and \(z^{(s)}_{\text{out}}\) denote the input and output feature maps at stage \(s\), \(v^{(s)}_0\) is the residual-convolved input, and the intermediate variables \(\hat z^{(s)}_1, z^{(s)}_1\), and \(\hat z^{(s)}_2\) represent successive outputs of attention and KAN blocks.  

The resulting hierarchical feature maps are passed through skip connections into the CNN-based decoder. 
The decoder consists of a sequence of upsampling stages, each composed of a deconvolution layer followed by convolutional blocks (Conv \(3\times3\) + BatchNorm + ReLU). At each stage, upsampled features are fused with encoder features via skip connections, restoring spatial resolution. A final \(1 \times 1\) convolution maps the decoded features to pixel-wise predictions.
 
\vspace{-5mm}
\subsection{Rational Base Functions}
We use rational functions as the base functions for KAN layers following KAT \cite{yang2024kolmogorovKAT}, rather than B-splines. 
Specifically, each edge function \(\phi(x)\) is parameterized as a rational over polynomials \(P(x), Q(x)\) of orders \(m\) and \(n\) respectively, using the Safe Padé Activation Unit (PAU) \cite{molina2019PAU}, as:  
{\small{
\begin{equation}
\label{equation:eq1}
\phi(x) = wF(x) = w \frac{P(x)}{1 + |Q(x)|},
\end{equation}
}}
where \(P(x) = a_0 + a_1x + \cdots + a_mx^m\), \(Q(x) = b_1x + \cdots + b_nx^n\), and \(w\) is a learnable scaling factor.  
This formulation avoids numerical instability (e.g., when \(Q(x) \to 0\), leading to \(\phi(x) \to \pm\infty\)) and provides a foundation for KAN layers that is computationally efficient (polynomial evaluation involves simple operations well-suited for parallel computing), expressive (capable of approximating a wider range of functions, including those with singularities or sharp variations), and stable during training. We empirically determined $m=3, \;n=4$ to be the best performing configuration.

\subsection{Group Rational KANs}

To enhance the feed-forward pathways, we replace the conventional MLP blocks (used in SwinUNETR) with Group Rational KANs (GR-KANs)
following the design of KAT \cite{yang2024kolmogorovKAT}. 
Instead of using fixed activations, KAN layers learn rational function expansions, enabling expressive non-linear mappings while retaining GPU efficiency. 
Group-KAN reduces parameter cost by sharing base functions across neurons. Note that a key limitation of the original KAN is that it requires a unique activation function 
for each input-output pair, leading to a parameter count of 
\(d_{in} \times d_{out}\) and high computational overhead. 
To overcome this, GR-KAN divides the input channels \(d_{in}\) into \(g\) groups, 
sharing rational function parameters within each group. 
This reduces the number of unique functions needed from \(d_{in} \times d_{out}\) to only \(g\), 
attaining significant savings in both parameters and FLOPs.  
Formally, given an input vector \(\mathbf{x}\), GR-KAN applies  
{\small{
\begin{equation}
GR\!-\!KAN(\mathbf{x}) = linear\;(group\_rational\;(\mathbf{x})),
\end{equation}
}}
where each group shares a rational function \(F(x)\) (see Eq.~(\ref{equation:eq1})) with shared polynomial coefficients \(\{a_i, b_j\}\), 
while each input–output edge retains its own learnable weight \(w\) in 
\(\phi(x) = wF(x)\).
This design preserves the expressive power of KANs while improving scalability and training efficiency, making it well-suited for data-scarce medical image segmentation tasks. 
Since we observed only minor performance differences across group sizes (1, 4, 8), we chose 8 for its better parallelization and faster runtime.

\section{Experimental Setup}
\label{sec:data}

\textbf{Datasets}: We evaluate our new method on 4 medical image segmentation datasets spanning different modalities: polyp tissue endoscopy (Kvasir-SEG \cite{jha2019kvasir}), skin lesion dermoscopy (ISIC 2017 \cite{codella2018isic}), abdominal CT (BCV 2016 \cite{landman2015bcv}), and cardiac MRI (MMWHS 2017 \cite{zhuang2019evaluationMMWHS}). The training sets were randomly sampled and contain 700 images, 2000 images, 24 volumes (1,982 2D slices), and 16 volumes (3,834 2D slices), respectively. For data efficacy experiments, we use 10\%, 25\%, 50\%, and 100\% of the 
training data. The corresponding test sets include randomly selected 300 images, 600 images, 6 volumes (603 2D slices), and 4 volumes (973 2D slices).

\noindent
\textbf{Baselines:}
We evaluate three CNN-based models (U-Net \cite{unet}, U-Net v2 \cite{peng2025unetv2}, UNeXt \cite{valanarasu2022unext}), two Transformer-based models (UNETR \cite{hatamizadeh2022unetr}, SwinUNETR \cite{tangSelfSupervisedPreTrainingSwin2022}), and one KAN-based model (UKAN \cite{li2025ukan}). We also construct a baseline, UKAT, by replacing the ViT \cite{vit} encoder in UNETR with the KAT encoder \cite{yang2024kolmogorovKAT}. All models are implemented in their smallest or comparable configurations to ensure fairness. For SwinUNETR and our proposed \mname{}, we additionally incorporate residual convolution blocks in the encoder stages as used in \cite{he2023swinunetrV2}.

\noindent
\textbf{Implementation Details:}
The implementations utilize 
and MONAI~\cite{cardoso2022monai}. The models were trained from scratch using the AdamW optimizer 
(initial learning rate = 0.0002, weight decay = 0.001) with a cosine annealing learning-rate schedule. The training used a combined Dice and cross-entropy loss on a single NVIDIA A10 GPU for 400 epochs with a batch size of 24. During training, $320 \times 320$ random crops were extracted and augmented with horizontal and vertical flips, random $90^{\circ}$ rotations, and Gaussian noise, with masks transformed identically. At inference, predictions were generated using overlapping patches with 50\% overlap.
\vspace{-3mm}
\section{Experimental Results}
\label{sec:results}

In this section, we present the results of different experiments with the baselines and our model.
\vspace{-2mm}
\subsection{Comparison with Known Architectures}
Our model, \mname, offers distinct advantages over existing architectures (see Table \ref{tab:main}).  Building on SwinUNETR, UKAST achieves consistent improvements, showing effective architectural enhancements with KANs. Compared to UKAN, another KAN-based model, UKAST achieves higher Dice scores on both 2D and 3D tasks. Most notably, UKAST excels on the 3D segmentation benchmarks, where it attains the best overall performance—outperforming all the baselines by a considerable margin on both BCV and MMWHS. Similarly, our spin on UNETR with KAN, UKAT (4.8M params.), outperforms the significantly larger UNETR (8.3M params.). 

\vspace{-3mm}
\subsection{Data and Computational Efficacy Study}
In the data efficacy experiments (see Table \ref{tab:data-efficacy}), UKAST demonstrates clear advantages over its baseline SwinUNETR across all training subsets, with particularly notable improvements on the 3D BCV dataset. This highlights UKAST’s ability to maintain robustness and performance even under limited data, enhancing ViT's ability to be competitive with CNNs in data-scarce settings. Similarly, UKAT consistently outperforms the much larger UNETR, showing that the incorporation of KAN layers effectively mitigates ViT's ineffectiveness with small training sets. 
\begin{table}
\centering
\resizebox{0.4\textwidth}{!}{
\begin{tabular}{c|lccc}
\hline
\hline
\textbf{Model} & \textbf{FFN} & \textbf{RC} & \textbf{GFLOPs} & \textbf{\# Params.} \\
\hline
\hline
SwinUNETR & MLP & \xmark & 1.2500 & 6,302,228 \\
\hgl{\mname{} (ours)}  & GR-KAN & \xmark  & 1.2467 & 6,302,836 \\
SwinUNETR & MLP  & \checkmark & 1.4419 & 7,183,508 \\
\hgl{\mname{} (ours)} & GR-KAN & \checkmark & 1.4386 & 7,184,116 \\
\hline
\hline
\end{tabular}
}
\caption{\small{\textbf{Model complexity comparison:} RC represents residual convolution in the encoder stage.}}
\label{tab:model_flops}
\end{table}
Additionally, we measured the model complexity by estimating FLOPs through a symbolic forward pass, summing multiply-accumulate operations (MACs) across all layers. As shown in Table~\ref{tab:model_flops}, replacing MLPs with GR-KANs yields lower FLOPs while incurring only a marginal increase in parameter count but providing better performance.
These results establish the potential of our proposed architecture as both parameter- and data-efficient that excels on the 
2D and 3D segmentation tasks.

\subsection{Ablation Study}

The ablation study (Table \ref{tab:ablation}) highlights four main points. 
(1) Replacing MLPs with GR-KANs in ViT encoders improves the performance by +3.5\% (+5.0\% in 3D). 
(2) Switching from ViT to SwinT yields a larger gain of +7.9\% (+9.0\%). 
(3) Within SwinT, replacing MLPs with GR-KANs provides an additional +2.1\% (+0.8\%). 
(4) When SwinT is augmented with residual convolution (RC), using GR-KANs instead of MLPs yields a further +0.4\% (+1.3\%).



\begin{table}[h]
\label{tab:kan_config}
\centering
\resizebox{0.8\linewidth}{!}{
\begin{tabular}{l l l  l l | c | c}
\hline
\hline
\multirow{2}{*}{\rotatebox{0}{\textbf{ViT}}} &
\multirow{2}{*}{\rotatebox{0}{\textbf{SwinT}}} &
\multirow{2}{*}{\rotatebox{0}{\textbf{RC}}} &
\multirow{2}{*}{\rotatebox{0}{\textbf{MLP}}} &
\multirow{2}{*}{\rotatebox{0}{\textbf{GR-KAN}}} &
\multicolumn{2}{c}{\textbf{Dice (\%)}} \\
\cline{6-7}
 & & & & & \textbf{2D} & \textbf{3D} \\
\hline
\hline
\checkmark &        & & \checkmark &                         & 69.2 & 61.5 \\
\checkmark &                       & &            &\checkmark  & 72.7 & 66.5 \\
\hline
 & \checkmark        &      & \checkmark                   & & 77.1 & 70.5 \\
   & \checkmark        &      &                    & \checkmark & 79.2 & 71.3 \\
 & \checkmark       & \checkmark       & \checkmark  & & 80.4 & 74.7 \\
  & \checkmark       & \checkmark       &   & \checkmark & 80.8 & 76.0 \\
 \hline
\hline
\end{tabular}
}
\caption{\small\textbf{Ablation results}:
Average Dice scores across two 2D datasets and two 3D datasets are reported.}
\label{ablation}
\label{tab:ablation}
\end{table}

\vspace{-7mm}
\section{Conclusions}
\label{sec:conclusion}

We introduced \mname{}, a novel Swin Transformer–based architecture enhanced with rational-function based KANs, addressing the inefficiencies of vanilla KANs and enabling Vision Transformers to excel in data-scarce medical image segmentation settings. 
\mname{} achieves state-of-the-art performance across four medical image segmentation benchmarks in diverse imaging modalities, with parameter counts comparable to SwinUNETR but taking fewer FLOPs.
Its robustness in limited annotation settings underscores its practicality for real-world medical image segmentation applications.

\section{Compliance with Ethical Standards}
This study was conducted using 4 publicly available datasets. No new human or animal data was collected. Ethical approval was not required, as confirmed by the licenses and terms of use associated with these open-access datasets.
\label{sec:compliance}

\small{
\bibliographystyle{IEEEbib}
\bibliography{strings,refs}
}
\end{document}